\pdfoutput=1

\documentclass[11pt]{article}

\newif\ifarxiv
\arxivtrue 

\ifarxiv
    \usepackage{acl}
\else
    \usepackage[review]{acl}
\fi
\usepackage{csquotes}
\usepackage{times}
\usepackage{latexsym}
\usepackage{amsmath}
\usepackage{enumitem}
\usepackage[T1]{fontenc}

\usepackage[utf8]{inputenc}

\usepackage{microtype}

\usepackage{inconsolata}
\usepackage{enumitem}
\usepackage{array}
\usepackage[capitalise]{cleveref}
\usepackage{listings}
\usepackage{graphicx}
\usepackage{longtable}
\usepackage{booktabs}
\usepackage{soul,xcolor}
\usepackage{longtable}
\sethlcolor{yellow}
\usepackage{subcaption}
\usepackage{multirow}
\usepackage{bm}
\usepackage{amsthm}
\usepackage{url}

 \usepackage[most]{tcolorbox}

\newtcolorbox{myquotebox}{
  colback=white!0, 
  colframe=black, 
  rounded corners,
  boxrule=0.5pt, 
  title=Prompt:,
  left=2mm, 
  right=2mm, 
  top=1mm, 
  bottom=1mm 
}
\usepackage{todonotes}
\usepackage{xspace}

\definecolor{lightgrey}{RGB}{158, 158, 158}
\definecolor{goldenrod}{rgb}{0,0,0.8}
\definecolor{deepred}{rgb}{0.6,0,0}
\definecolor{deepgreen}{rgb}{0,0.5,0}
\definecolor{pink}{RGB}{219, 48, 122}
\definecolor{forestgreen}{RGB}{34,139,34}
\definecolor{goldenrod}{RGB}{218,165,32}
\definecolor{sepia}{RGB}{112,66,20}

\crefname{figure}{Fig.}{Figs.}
\crefname{table}{Table}{Tables}
\crefname{App.}{App.}{App.}
\crefname{section}{§}{§§}
\crefname{equation}{Eq.}{Eqs.}

\newcommand\myparagraph[1]{
\vskip 0.05in 
\noindent{\bf {#1}}}
\newcommand*\samethanks[1][\value{footnote}]{\footnotemark[#1]}

\title{pdfQA: Diverse, Challenging, and Realistic Question Answering over PDFs}

\author{ 
    Tobias Schimanski\textsuperscript{\rm 1},
    Imene Kolli\textsuperscript{\rm 1}, 
    Yu Fan\textsuperscript{\rm 2}, 
    Ario Saeid Vaghefi\textsuperscript{\rm 1},\\
    \textbf{Jingwei Ni}\textsuperscript{\rm 2}\thanks{Equal Supervision.},
    \textbf{Elliott Ash}\textsuperscript{\rm 2}\samethanks,
    \textbf{Markus Leippold}\textsuperscript{\rm 1,3}\samethanks  \\
    \textsuperscript{\rm 1}University of Zurich \hspace{5mm}
    \textsuperscript{\rm 2}ETH Zurich \hspace{5mm} 
    \textsuperscript{\rm 3}Swiss Finance Institute (SFI) \\
    \texttt{\{tobias.schimanski, imene.kolli, markus.leippold\}@df.uzh.ch} \\
    \texttt{\{jingni, yufan, ashe\}@ethz.ch}, \texttt{saeid.vaghefi@geo.uzh.ch}
}

\begin{document}
\maketitle
\begin{abstract}
PDFs are the second-most used document type on the internet (after HTML). Yet, existing QA datasets commonly start from text sources or only address specific domains. In this paper, we present \textit{pdfQA}, a multi-domain 2K human-annotated (\textit{real-pdfQA}) and 2K synthetic dataset (\textit{syn-pdfQA}) differentiating QA pairs in ten complexity dimensions (e.g., file type, source modality, source position, answer type). We apply and evaluate quality and difficulty filters on both datasets, obtaining valid and challenging QA pairs. We answer the questions with open-source LLMs, revealing existing challenges that correlate with our complexity dimensions. \textit{pdfQA} presents a basis for end-to-end QA pipeline evaluation, testing diverse skill sets and local optimizations (e.g., in information retrieval or parsing).
\ifarxiv
\footnote{Data: \url{https://github.com/tobischimanski/pdfQA}.}
\else
\footnote{Core data in submission. All data will be open-sourced.}
\fi
\end{abstract}

\section{Introduction} \label{sec:introduction}
Large Language Models (LLMs) are used for a multitude of tasks, showing increasingly impressive capabilities in coding, math, and general QA benchmarks \citep[e.g.,][]{openai2024gpt4technicalreport, qwen3, grattafiori2024llama3herdmodels}. One major area of usage is evidence-based question answering (EBQA), nowadays most commonly in generative search engines or retrieval augmented generation (RAG). However, there exists an ongoing debate about the reliability of EBQA \citep{liu2023evaluatingverifiabilitygenerativesearch}, with pronounced problems outlined in open-source models \citep{schimanski-etal-2024-towards}. One major issue in measuring performance, and therefore improving models, is the lack of realistic benchmarks \citep{bean2025measuringmattersconstructvalidity}.

A starting point for realistic QA is handling different document types. Yet, the most common evidence-based QA datasets start from preprocessed text samples \citep[e.g.,][]{rajpurkar2016squad100000questionsmachine, kwiatkowski-etal-2019-natural, yang2018hotpotqadatasetdiverseexplainable}. In this light, PDFs are largely underexplored. Even within benchmarks addressing PDFs, there remain three shortcomings. First, most datasets are concentrated on individual use cases, such as scientific reports \citep{singh2024scidqadeepreadingcomprehension, xie2025pdfwukonglargemultimodalmodel} or financial reporting \citep{chen2022finqadatasetnumericalreasoning}. Second, even in more comprehensive benchmarks \citep{hui2024udabenchmarksuiteretrieval}, there prevails a lack of differentiation regarding the complexity dimensions of questions (e.g., text or table as a source, source quantity, or question difficulty). Commonly, the complexity dimensions are homogeneous within one benchmark (e.g., only questions on tables). Third, the quality and difficulty of questions and answers are commonly not assessed. High quality is frequently just assumed, especially for human-annotated data, and despite prior work outlining issues \citep{liu2024ecbdevidencecenteredbenchmarkdesign, klie2022annotationerrordetectionanalyzing, calamai-etal-2025-benchmarking}. 

To address these shortcomings, we propose \textit{pdfQA}, a 4K dataset containing 2K synthetic (\textit{syn-pdfQA}) and 2K human-annotated QA pairs over PDFs (\textit{real-pdfQA}). In \textit{syn-pdfQA}, we leverage a synthetic data generation process to create QA pairs of ten complexity dimensions (e.g., file types, question and answer types, or spread of relevant sources across a document). In \textit{real-pdfQA}, we combine nine existing human-annotated benchmarks on QA over PDFs, and search and prepare the source PDF documents. Both datasets undergo rigorous quality and difficulty filters.

Then, we answer the questions in \textit{pdfQA} with open-source models. On \textit{syn-pdfQA}, we find that our complexity dimensions indeed correlate with model performance, outlining analysis potentials. We also find that the human-annotated QA pairs in \textit{real-pdfQA} are more challenging. In this way, \textit{pdfQA} contains QA pairs on diverse source documents and relevant sources, classified by a QA pair taxonomy, as well as challenging real-world tasks. Hence, \textit{pdfQA} can support various (stepwise) end-to-end evaluation scenarios (e.g., in parsing).

\begin{figure*}[t]
    \centering
	\includegraphics[width=0.8\textwidth]{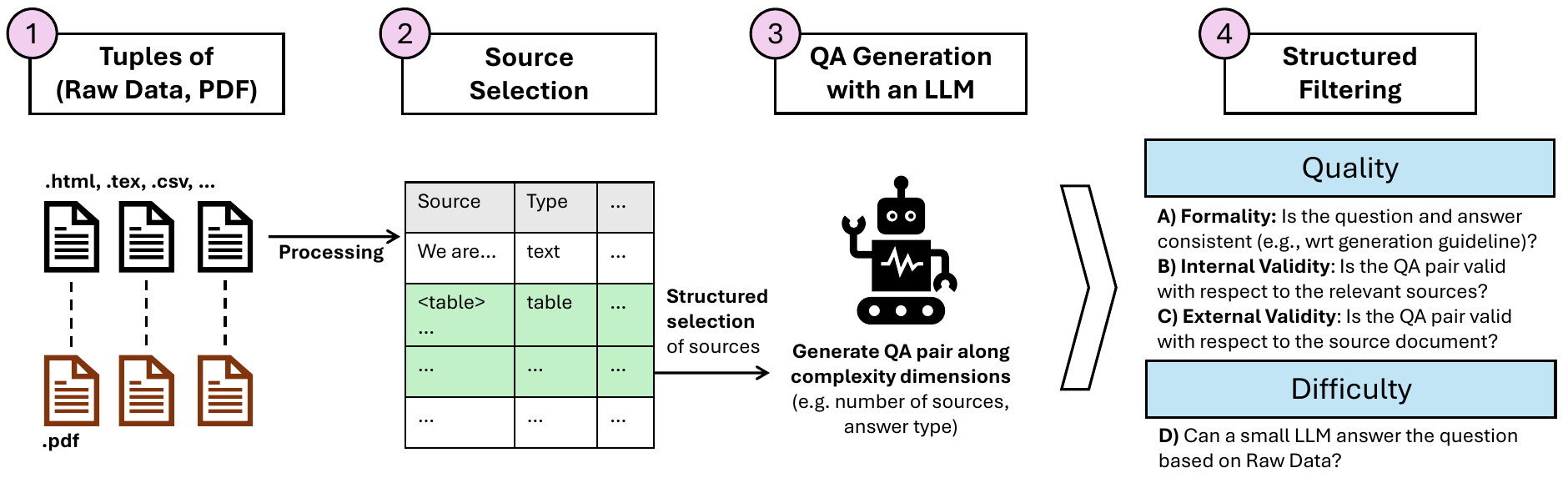}
	\caption{\textbf{Synthetic Data Generation Pipeline}. The input is tuples of PDFs and the same document in a structured format (e.g., .tex or .html files). We use financial reports, research articles, books, and sustainability disclosures (1). These are processed into structured sources (2), upon which an LLM creates QA pairs with different complexity dimensions (3). We apply quality and difficulty filters to ensure valid and challenging QA pairs in \textit{pdfQA} (4).}
	\label{fig:overview}
\vspace{-0.5em}
\end{figure*}

\section{pdfQA}
\textit{pdfQA} consists of two types of datasets: \textit{syn-pdfQA}, which uses an LLM-based pipeline to create diverse QA pairs of different complexity dimensions, and \textit{real-pdfQA}, a collection of nine domain-specific, human-annotated datasets on PDFs. Both datasets are filtered to ensure the quality and difficulty of QA pairs.

\subsection{syn-pdfQA}
To create \textit{syn-pdfQA}, we proceed in four steps (see Figure \ref{fig:overview}). First, we search for PDFs that are accompanied with structured raw data files, such as \enquote{\textit{.tex}} or \enquote{\textit{.html}} files. This is crucial because it allows us to create QA pairs upon the raw files (which are a perfectly parsed version of the PDF), while the final benchmark can start from the PDF file. We sample a diverse set of input files from: financial reporting, research articles, books, and sustainability disclosures (see App. \ref{app:input_data_syn} for details). 

Second, we process the raw files into a structured format that separates individual sources on a paragraph-level, labels them as text or table, and traces characteristics like position within the document or length of the document. Besides, we cluster all sources with an embedding model to not only track positional, but also semantic relations within the sources (for details, see App. \ref{app:embeddings_clustering}). For the generation of every QA pair, we choose a random seed paragraph and a set of 5-15 sources that are either around the seed paragraph (positional proximity) or in the same cluster as the seed paragraph (semantic proximity).

Third, we use these sampled sources and instruct GPT-4.1 to create a question and answer upon them (for prompt setup, see App. \ref{app:qa_generation_process}). While creating QA pairs, we specifically ask the model to create certain types of questions and answers. Combined with the structural knowledge about the raw data, we can create various complexity dimensions (see also App. \ref{app:complexity_dimensions}). We can differentiate dimensions created based on structural knowledge (sn) and based on the LLM (llm):
\begin{itemize}[nosep]
    \item \textit{answer type (llm)}: yes/no, value extraction, single word, or open-ended answers,
    \item \textit{answer length (sn)}: the length of the answer,
    \item \textit{reasoning (llm)}: whether the answer is a replication of information or needs reasoning, 
    \item \textit{question difficulty (llm)}: a pre-defined level of difficulty (simple, medium, hard),
    \item \textit{modalities (sn)}: the modalities used to answer the question (e.g., text, tables, mixed modalities), 
    \item \textit{number of relevant sources (sn)}: how many relevant sources are needed to answer the question, 
    \item \textit{source spread (sn)}: a distance measure of how much text is between the first and the last relevant source, 
    \item \textit{source position (sn)}: a proxy for where in the file the relevant sources are clustered, 
    \item \textit{file type (sn)}: an indication of which file type the question belongs to,
    \item \textit{file length (sn)}: the length of the file.
\end{itemize}

Fourth, we apply quality and difficulty filters to the generated QA pairs (see step 4 in Figure \ref{fig:overview}). The quality filter checks three dimensions with the help of GPT-4.1-mini. First, we determine whether the QA generation followed the formal criteria defined in the generation process. Specifically, we control whether the question refers to the entire document (e.g., no question stating to refer to \enquote{the table above}), the answer is deterministic (e.g., no question that says \enquote{name \textit{some} of the criteria}), and the answer type is correct (e.g., a yes/no answer type contains indeed only a \enquote{yes} or \enquote{no}). Second, we check internal validity. This means, given the relevant sources and the question, is the answer correct? Third, we check external validity. This means, given the relevant sources and the top-k semantically related sources in the raw data, is the answer still correct? We check this because during the generation process, there might be missing pieces of information in the selected sources that could contradict the answer (see App. \ref{app:quality_filters}).

Besides the quality filters, the final dataset is also filtered for more challenging questions. We apply a simple heuristic: if GPT-4o-mini can answer the question given the entire raw data file, then it is too simple. The final dataset only contains those QA pairs where the model failed to answer (see App. \ref{app:difficulty_filter}). We observe that for the synthetically generated QA pairs, 20.4\% are removed by the quality filters, and out of the remaining, 67.5\% are filtered for not being difficult enough (see App. \ref{app:filters_on_syn}). Overall, we produce 7655 input QA pairs with our pipeline, of which 1982 remain after filtering (Table \ref{tab:filtering_results}). Despite our rigorous quality filtering approaches, the final dataset could contain erroneous QA pairs. To mitigate these concerns, we employ two human annotators to validate 200 filtered QA pairs. We find 88\% are entirely correct with an annotator agreement of 93\% (see App. \ref{app:hand_annotation}), strengthening our confidence that we created a correct QA dataset.

\begin{table}[t]
\centering
\begin{tabular}{ccc}
\hline
                                                                       & syn-pdfQA                                           & real-pdfQA                                           \\ \hline
Input QA pairs                                                         & 7,655                                               & 22,866                                               \\
After filtering                                                        & 1,982                                               & 2,041                                                \\ \hline
\begin{tabular}[c]{@{}c@{}}Hand annotation\\ (n filtered QA pairs)\end{tabular} & \begin{tabular}[c]{@{}c@{}}88\%\\ (200)\end{tabular} & \begin{tabular}[c]{@{}c@{}}91\%\\ (100)\end{tabular} \\ \hline
\end{tabular}
\caption{\textbf{Final datasets after filtering}. The input for \textit{syn-pdfQA} is synthetically generated QA pairs, while for \textit{real-pdfQA}, it is QA pairs from nine benchmarks. Both are filtered for quality and difficulty. Then, n filtered QA pairs are hand-annotated, checking whether they are correct.}
\label{tab:filtering_results}
\end{table}

\subsection{real-pdfQA}
To create \textit{real-pdfQA}, we proceed in two steps. First, we search for human-annotated QA datasets that start from PDF files. Although many datasets start from PDF theoretically, the files themselves are commonly not open-sourced. Thus, we also search for the corresponding PDFs (similar to \citealp{hui2024udabenchmarksuiteretrieval}). Besides, we also filter for datasets that complement a QA pair with source information, making it compatible with our later filtering steps. We use nine human-annotated QA datasets that start from PDFs, including FinQA \citep{chen2022finqadatasetnumericalreasoning}, FinanceBench \citep{islam2023financebenchnewbenchmarkfinancial}, ClimateFinanceBench \citep{mankour2025climatefinancebench}, ClimRetrieve \citep{schimanski-etal-2024-climretrieve}, FeTaQA \citep{nan-etal-2022-fetaqa}, NaturalQuestions \citep{kwiatkowski-etal-2019-natural}, PaperTab and PaperText in Qasper \citep{dasigi2021datasetinformationseekingquestionsanswers}, and Tat-QA \citep{zhu2021tatqaquestionansweringbenchmark} (for the last five datasets, we make use of the PDF collection in \citealp{hui2024udabenchmarksuiteretrieval}).

Second, we apply our quality and difficulty filters to these datasets (step 4 in Figure \ref{fig:overview}). While the datasets are human-annotated, we view the quality control as a response to concerns about NLP benchmark validity \citep[e.g.,][]{McIntosh_2025, calamai-etal-2025-benchmarking, subramonian-etal-2023-takes} as well as adjusting the dataset to our specific use case needs \citep{bean2025measuringmattersconstructvalidity}. Indeed, we find that our quality filters eliminate 58\% of the QA pairs. The major reason for filtering out datapoints is their lacking generalization towards an entire document. This stems from their original use cases. A large set of questions refers to one specific text or table (e.g., "What is the value in \textit{this} table?"). With our quality filters, we aim to ensure validity with respect to the entire document, thereby filtering these QA pairs out (see App. \ref{app:filters_on_real}). Among the quality-filtered QA pairs, 78\% are filtered out by our difficulty filter. This means GPT-4o-mini was able to answer the question correctly, given the entire document. Hence, starting with 22K entries, our filtered real-pdfQA contains 2041 QA pairs (Table \ref{tab:filtering_results}). Once again, we hand-annotate whether these QA pairs are indeed valid, finding 91\% correctness among randomly sampled 100 QA pairs (see App. \ref{app:hand_annotation}).

\section{Experiments}

\begin{table}[t]
\centering
\small
\begin{tabular}{llcc}
\toprule
Dimension & Subdimension
& gpt-oss-120b
& llama3-8b \\
\midrule
\multirow{4}{*}{file type} 
& financial 
& 4.266 & 3.288 \\
& research 
& 4.625 & 3.871 \\
& sustainability
& 4.104 & 3.010 \\
& books
& 4.307 & 3.246 \\
\midrule
\multirow{2}{*}{reasoning} 
& reasoning 
& 4.323 & 3.363 \\
& replicate 
& 4.479 & 3.641 \\
\midrule
\multirow{3}{*}{modalities} 
& multimodal 
& 4.495 & 3.446 \\
& table 
& 4.177 & 3.123 \\
& text 
& 4.538 & 3.840 \\
\midrule
\multirow{2}{*}{file length} 
& < median 
& 4.560 & 3.788 \\
& > median 
& 4.234 & 3.203 \\
\bottomrule
\end{tabular}
\caption{\textbf{Model performance on selected complexity dimensions of \textit{syn-pdfQA}}. Correctness scores are between 1-5 (Figure \ref{fig:answer_quality_prompt}). Full results are in Table \ref{tab:full_result_on_syn}.}
\label{tab:selected_result_on_syn}
\vspace{-0.25em}
\end{table}

This section shows baseline results and analysis potentials for \textit{pdfQA}. For this purpose, we develop a simple QA process starting from the source PDF documents. We parse the PDFs using the \textit{PyMUPDF} parser\footnote{\url{https://pymupdf.readthedocs.io/en/latest/}}, and input the entire parsed text as a context and the corresponding question into the prompt template in Figure \ref{fig:prompt_qa_pair_gen}. We answer the questions with five open-source models: gpt-oss-120b and gpt-oss-20b \citep{openai2025gptoss120bgptoss20bmodel}, qwen3-next \citep{qwen3}, gemma3-27b \citep{gemmateam2025gemma3technicalreport}, and llama3-8b \citep{grattafiori2024llama3herdmodels}.
Then, we use the G-Eval correctness prompt in Figure \ref{fig:answer_quality_prompt} to evaluate the answer. This prompt scales the answer between 1-5, where 5 indicates a completely correct answer (see App. \ref{app:details_on_model_setup}).

For \textit{syn-pdfQA}, we can differentiate the results into all complexity dimensions (see Table \ref{tab:selected_result_on_syn}, \ref{tab:full_result_on_syn}). Generally, as expected, rising model size increases performance. QA on research articles seems the easiest. The models consistently perform worse on value-related questions (e.g., asking for table value interpretations). This coincides with results on different modalities, where questions on tables are more challenging than those based on text. Similarly, scores are lower when the answer requires more reasoning, when there are more relevant sources, when the file is longer, and when the relevant sources are more widely spread across the document. We also observe the tendency that if relevant sources cluster in the middle of the document, the model fails more often \citep[see also e.g.,][]{liu2023lostmiddlelanguagemodels}. Increasing model size decreases the intensity of these observations.

\begin{figure}[t]
    \centering
	\includegraphics[width=0.5\textwidth]{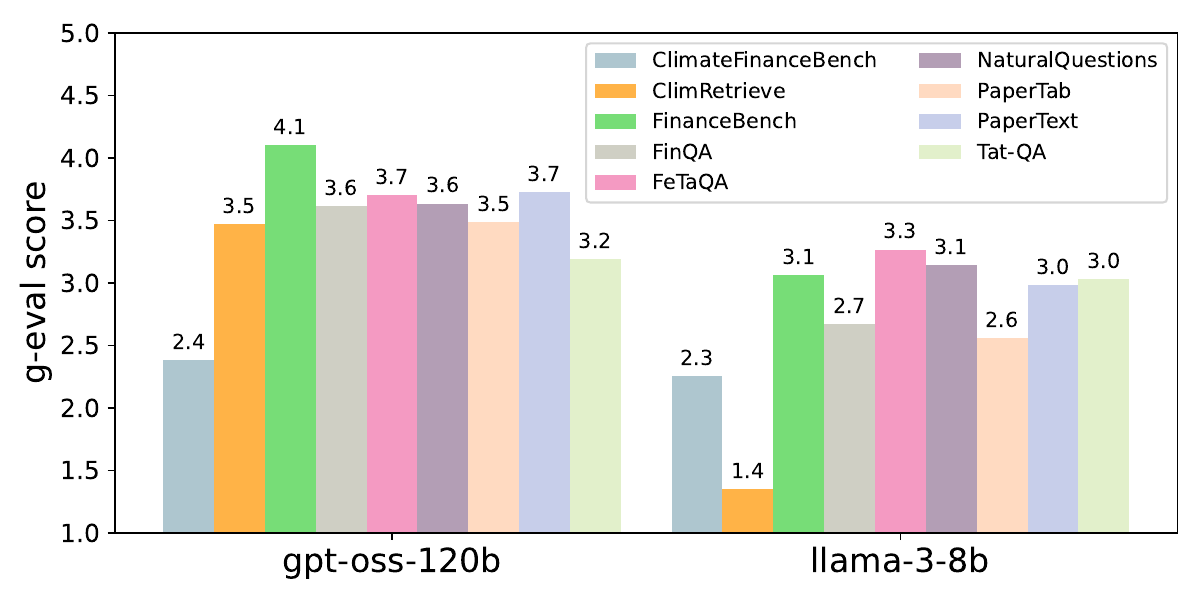}
	\caption{\textbf{Model performance on datasets in \textit{real-pdfQA}}. Correctness scores are between 1-5 (Figure \ref{fig:answer_quality_prompt}).}
	\label{fig:results_on_real_reduced}
\vspace{-0.25em}
\end{figure}

For \textit{real-pdfQA}, we can only observe the results for different datasets (see Figure \ref{fig:results_on_real_reduced}, Table \ref{tab:full_result_on_real}). However, we find significant value in using human-annotated QA pairs. While \textit{syn-pdfQA} allows diverse investigations along complexity dimensions, \textit{real-pdfQA} contains more challenging QA pairs, with average model scores consistently lower than in \textit{syn-pdfQA}. Among the most complicated datasets are those that require intense domain knowledge (ClimRetrieve, ClimateFinanceBench) and advanced reasoning (Tat-QA).

Collectively, these insights align with intuition, strengthening our belief in the validity of the data. At the same time, these results can serve as a starting point for several optimizations of QA approaches on PDFs.

\section{Literature Background}
There are two major streams of related work. First, prior work has already focused on single components of PDFs, mostly through the lens of visual (language) models or multimodal models. This includes answering question over single figures and charts \citep{kahou2018figureqaannotatedfiguredataset,methani2020plotqareasoningscientificplots,masry2022chartqabenchmarkquestionanswering}, tables \citep{zheng2024multimodaltableunderstanding}, or in multimodal settings \citep{pramanick2025spiqadatasetmultimodalquestion, Yihao2024mmvqa}. We contemplate PDFs as a whole, asking questions over an entire document.

Second, prior work has explored single domains in question answering over PDFs, such as financial reporting \citep{mankour2025climatefinancebench, zhu2021tatqaquestionansweringbenchmark}, research papers \citep{xia2024docgenomeopenlargescalescientific, pramanick2025spiqadatasetmultimodalquestion, xie2025pdfwukonglargemultimodalmodel}, or sustainability reporting \citep{ni2023chatreportdemocratizingsustainabilitydisclosure, gehricke2025to}. We integrate all human-annotated datasets into \textit{pdfQA}, if PDF sources are available.\footnote{We encourge researchers to reach out to us if they want to include their datasets in \textit{pdfQA}.} Closest to our work is UDA \citep{hui2024udabenchmarksuiteretrieval}. It also compiles a human-annotated PDF-based QA dataset. We integrate UDA, extend it considerably, and apply our filtering steps, eliminating ca. 90\% of the original data.

\section{Conclusion}
In this paper, we introduce \textit{pdfQA}, a dataset comprising diverse synthetic as well as challenging human-annotated QA pairs based on PDFs. We rigorously filter for quality and validity. By running a simple long context experiment, we show the analysis potential of the dataset. \textit{pdfQA} can fill a critical gap in obtaining insights on local improvements in end-to-end pipelines regarding various complexity dimensions, and process steps (e.g., information retrieval on PDFs). Hence, it opens up new venues for researchers and practitioners interested in realistic QA.

\section*{Limitations}
As with all projects, \textit{pdfQA} has limitations. First, even after rigorously filtering and hand-verifying the dataset, it may still contain error cases. Although our tests signal a high quality, follow-up research can improve the filtering steps. Specifically, we cannot fully account for external validity, i.e., whether the QA pairs are truly valid with respect to the entire document. We try to proxy for that with a top-k approach. 

Second, our quality filters seem to be conservative in deeming QA pairs as valid (also see App. \ref{app:filters_on_syn} and \ref{app:filters_on_real}). Thus, we refrain from making claims about the quality of the investigated datasets. Other work has and should more cleanly identify actual quality issues in QA datasets \citep[e.g.,][]{liu2024ecbdevidencecenteredbenchmarkdesign, klie2022annotationerrordetectionanalyzing, calamai-etal-2025-benchmarking}. The focus of this paper lies on producing challenging and valid, however possibly over-conservatively filtered QA pairs. 

Third, the experiments in this paper only provide evidence on challenges of long-context question answering over parsed PDFs. While the focus of the paper does not lie on exploring sophisticated solutions, we cannot make statements about whether approaches like retrieval augmented generation already or visual models deliver a different picture. Future research should focus on exploring complexity dimensions for these approaches.

\section*{Ethics Statement}
\myparagraph{Human Annotation}: In this work, all human annotators are Graduate or doctoral researchers who have good knowledge about scientific communication and entailment. They are officially hired and have full knowledge of the context and utility of the collected data. We adhered strictly to ethical guidelines, respecting the dignity, rights, safety, and well-being of all participants. 

\myparagraph{Data Privacy or Bias}: There are no data privacy issues or biases against certain demographics with regard to the data collected from real-world applications and LLM generations. All artifacts we use are under a Creative Commons license. We also notice no ethical risks associated with this work

\myparagraph{Reproducibility Statement}: To ensure full reproducibility, we will disclose all codes and data used in this project, as well as the LLM generations, GPT, and human annotations. For all API-based models, we always fix the temperature to 0.

\ifarxiv
\section*{Acknowledgements} 
This paper has received funding from the Swiss
National Science Foundation (SNSF) under the project `How sustainable is sustainable finance? Impact evaluation and automated greenwashing detection' (Grant Agreement No. 100018\_207800).
\else
\fi

\bibliography{custom}
\appendix

\section{Input data for synthetic data generation pipeline} \label{app:input_data_syn}
For our synthetic data pipeline, we use four file types: financial reporting, research articles, books, and sustainability disclosures. For financial reporting, we use the annual reports (Form 10-K documents) of US firms. We obtain them from the Securities Exchange Commission (SEC) Electronic Data Gathering, Analysis, and Retrieval (EDGAR) platform. We obtain an ".htm" file as raw data and the PDF. For research articles, we use computer science research articles on Arxiv that have more than 5 citations. The citation count should ensure that the articles meet a minimal threshold of quality, assuming that non-cited articles might have lower quality. We obtain a ".tex" file as raw data and the PDF. For books, we use open-sourced books on Springer Nature. 
We obtain a ".txt" file as raw data and the PDF. For sustainability disclosure, we use documents prepared by the Nexxar\footnote{Nexxar is a startup preparing sustainability reports in both HTML and PDF form to make them more accessible to online tools, see also \url{https://nexxar.com/who-we-are.html}.}. We obtain a ".csv" file as raw data containing all tables in the reports and the full PDF. This means, on sustainability reports, we only ask table-related questions.


\section{Clustering sources with embeddings}\label{app:embeddings_clustering}
Once the raw data is structured into a paragraph format, we want to cluster it with an embedding model to obtain semantically, not only positionally connected paragraphs. For this, we use OpenAI's \enquote{text-embedding-3-small} model. We create the number of clusters by dividing the number of paragraphs in the report by 15.

\section{LLM prompt for the QA generation process} \label{app:qa_generation_process}
The prompt for generating the QA pair follows the prompt template in Figure \ref{fig:prompt_qa_pair_gen}. The template uses a set of sources that are obtained either arbitrarily around a random seed source or out of a semantic cluster of sources. The number of sources inputted is randomized between 5-15. The prompt also obtains a random set of guidelines that contains one of the options in each category of \textit{source quantity}, \textit{answer type}, \textit{replicate or reasoning}, \textit{modality}, and \textit{difficulty}, as listed in Figure \ref{fig:guidelines_qa_pair_gen}. We create 50 questions per file (that are then input into the quality and difficulty filters). For QA pair generation, we use OpenAI's \enquote{gpt-4.1-2025-04-14} with \textit{temperature=0}.

\begin{figure}[ht]
\begin{lstlisting}[frame=single, basicstyle=\ttfamily\scriptsize, xleftmargin=0pt, numbers=none]
You are a domain expert in {domain} and are provided with SOURCES from a domain document. Your task is to create a QUESTION and ANSWER based on the SOURCES.


Use the following questions SOURCES:
[Begin of SOURCES]
{sources}
[End of SOURCES]


Create a QUESTION and an ANSWER for the QUESTION following these GUIDELINES:
{guidelines}

Your task is to create a QUESTION that is realistic in the context of a question-answering benchmark over reports. Do not come up with artificial questions that would never appear in reality. In reality, the provided sources appear throughout the entire document. Thus, do not refer to a source identifier or assume that the source is standalone. The QUESTION should be designed to make sense with respect to the entire document (that you have no full access to).

Please output the answer in a JSON format using the keys "question", "answer", and "sources". "question" contains the QUESTION you have produced. "answer" contains the ANSWER to the question. "sources" includes the source identifiers in the SOURCES (e.g., ["Source 1", "Source 2"]). Strictly cite only the sources that were actually used to create the question, not necessarily all the sources given.

Output the JSON file with the keys "question", "answer", and "sources":
"""
\end{lstlisting}
\caption{Prompt for generating QA pairs. \enquote{sources} is replaced with a set of 5-15 sources from the underlying file, \enquote{guidelines} is replaced with one of the options in each category of \textit{source quantity}, \textit{answer type}, \textit{replicate or reasoning}, \textit{modality}, and \textit{difficulty}, as listed in Figure \ref{fig:guidelines_qa_pair_gen}.}
\label{fig:prompt_qa_pair_gen}
\end{figure}

\begin{figure}[ht]
\begin{lstlisting}[frame=single, basicstyle=\ttfamily\scriptsize, xleftmargin=0pt, numbers=none]
# source quantity
"arbitrary sources": "The QUESTION must be answerable using one or more given SOURCES. Strive for connecting SOURCES in a logical manner.",
"strict multiple sources": "The QUESTION must be answerable using as many as possible given SOURCES. However, produce a single QUESTION that only uses logically connected SOURCES.",

# answer type
"yes-no-question": "The QUESTION must be answerable by strictly only a 'Yes' or 'No'. Please do not use any other words in the answer.",
"value-question": "The QUESTION must be answerable by a single value.",
"word-answer": "The QUESTION must be answerable by one to five words, without forming a full sentence.",
"one-sentence-answer": "The QUESTION must be answerable by a single sentence.",
"open-ended-question-short": "The QUESTION must be answerable by a short open-ended answer.",
"open-ended-question-long": "The QUESTION must be answerable by a long open-ended answer.",

# replicate or reasoning
"replicate": "The QUESTION must be answerable by a simple replication of parts of the SOURCES or a simple summary of SOURCES.",
"reasoning": "The QUESTION must be answerable by a reasoning process over the SOURCES.",

# modality
"text-only": "The QUESTION must be answerable with only the contents of a text.",
"table-only": "The QUESTION must be answerable with only the contents of a table, including its caption.",
"mixed-modality": "The QUESTION must be answerable through a combination of multiple modalities between text, table, or others.",
"clustering": "The QUESTION can make use of a combination of multiple modalities between text, table, or a others.",

# difficulty
"simple": "The QUESTION is very simple and straightforward to answer given the SOURCES.",
"medium": "The QUESTION is moderately complex and requires some reasoning to answer, given the SOURCES.",
"complex": "The QUESTION is very complex and requires a lot of reasoning to answe,r given the SOURCES."
\end{lstlisting}
\caption{Options for guidelines.}
\label{fig:guidelines_qa_pair_gen}
\end{figure}

\section{QA Complexity Dimensions}\label{app:complexity_dimensions}
The generation process described in App. \ref{app:qa_generation_process} cumulates in QA pairs that have the complexity dimensions described in Table \ref{tab:complexity_dimensions}.

\begin{table*}[t]
\centering
\small
\begin{tabular}{p{0.27\textwidth} p{0.68\textwidth}}
\hline
Complexity Dimension & Explanation                                                                                                                                            \\ \hline
Answer Type          & format of the answer, including yes/no answers, value answers, answers of single words, answers of single sentences, short and long open-ended answers \\
Answer Length        & length of the answer (extension/complement to "Answer Type")                                                                                           \\
Reasoning            & the question requires reasoning or mere replication of the information in the texts                                                                    \\
Question Difficulty  & simple, medium or complex question requiring differing levels of reasoning (extension/complement of "Reasoning")                                       \\
Modalities           & text-based, table-based or mixed modality question                                                                                                     \\
Number of Sources    & number of relevant sources (paragraphs) needed to answer the question                                                                                  \\
Source Spread        & number of words between first and last relevant source; as a proxy for dispersion of relevant sources                                                  \\
Sources Position     & location of the majority of sources given by their location in the first 25\% (beginning), 25-50\% (middle 1), 50-75\% (middle 2), 75-100\% (end) of the sources                         \\
File Type            & financial reporting, research article, book, or sustainability disclosure    
       \\
File Length            & word count of the raw file    \\ \hline
\end{tabular}
\caption{Complexity Dimensions. This table shows the different complexity dimensions induced by the QA generation process.}
\label{tab:complexity_dimensions}
\end{table*}

\section{Quality Filters}\label{app:quality_filters}
Figure \ref{fig:formality_filter} shows the formality filtering for the QA pairs, ensuring that the question generation process was successful with respect to formal criteria, like whether the question is consistent with respect to an entire document, whether the answer is deterministic, or whether the guidelines in creation were fulfilled. Figure \ref{fig:validity_filter} shows the prompt for the validity filter for the QA pairs, inspired by G-Eval \citep{liu2023gevalnlgevaluationusing}. This filter should ensure that the answers to the questions are indeed valid with respect to the sources. We run two setups. \textit{Internal validity}, in which the \enquote{Sources} in the prompt are those that the model has seen during the generation of the question. And \textit{External validity}, in which the \enquote{Sources} in the prompt are extended by a top-k with k=5 semantic search with the question. This should aim for detecting whether there are contrasting sources regarding the question and answer that would falsify the validity. We use OpenAI's \enquote{gpt-4.1-mini-2025-04-14} to answer the prompts and \enquote{text-embedding-3-small} for the semantic search.

\begin{figure}[ht]
\begin{lstlisting}[frame=single, basicstyle=\ttfamily\scriptsize, xleftmargin=0pt, numbers=none]
"""You will be given a question, a guideline, and an answer.

Your task is to check whether the formal criteria of the question and answer are fulfilled.
The evaluation is based on the following criteria. All must be satisfied for the final result to be "yes".

Evaluation Criteria:
- Is the question free of references that could be misunderstood in the context of a full document? (e.g., the question does not refer vaguely to "the table")
- Is the question unambiguous and likely to have a deterministic answer? (disallowed are questions such as "What is one example for...?")
- Does the question align with the guideline? (do only judge clear misalignments, e.g. a yes/no question was not answered with "yes" or "no")

Question:
----------
{Question}
----------

Guideline:
----------
{Guideline}
----------

Answer:
----------
{Answer}
----------

Instruction:
Answer only with "yes" or "no".
If all evaluation criteria are met, respond with "yes". Otherwise, respond with "no".

Your answer:
"""
\end{lstlisting}
\caption{Formality filter for QA quality. \enquote{Question} is replaced with the question under investigation, \enquote{Answer} with the answer under investigation. \enquote{Guideline} is replaced with its corresponding \textit{answer type} as defined in Figure \ref{fig:guidelines_qa_pair_gen}.}
\label{fig:formality_filter}
\end{figure}

\begin{figure}[ht]
\begin{lstlisting}[frame=single, basicstyle=\ttfamily\scriptsize, xleftmargin=0pt, numbers=none]
"""You will be given a set of sources, a question and an answer.

Your task is to rate the faithfulness of the answer. Please make sure you read and understand these instructions carefully.

Evaluation Criteria:
Faithfulness (1-5) - this evaluates whether the generated answer strictly adheres to the set of sources without introducing unsupported or contradicting claims.

Evaluation Steps:
1. Read the sources and question: Identify relevant information in the sources that directly addresses the question.
2. Compare the answer to the sources: Check if the answer strictly adheres to the sources and avoids unsupported or contradicting claims.
3. Assign a faithfulness score (1-5):
- 1: The answer misrepresents or contradicts the sources.
- 2: Contains multiple unsupported claims, errors, or contradictions.
- 3: Partially faithful with some unsupported claims or inaccuracies.
- 4: Mostly faithful with minor deviations.
- 5: Fully faithful and strictly adheres to the sources.

Set of sources:
----------
{Sources}
----------

Question:
----------
{Question}
----------

Answer:
----------
{Answer}
----------

Evaluation Form (ouput ONLY a single score - nothing else):
- Faithfulness:
"""
\end{lstlisting}
\caption{Validity filter for QA quality. \enquote{Question} and \enquote{Answer} are the question and answer under investigation. \enquote{Sources} is replaced with a set of sources that refer to the question.}
\label{fig:validity_filter}
\end{figure}

\section{Difficulty Filter}\label{app:difficulty_filter}
For all QA pairs passing the quality filters, we apply an additional difficulty filter. This filter should simply eliminate QA pairs that are too easy. For doing this, we take the raw data document (i.e., the perfectly parsed version, see Figure \ref{fig:overview}) and plug it into its entirety into the question answering prompt as defined in Figure \ref{fig:answer_generation_prompt}. Then, we evaluate the answer with another G-Eval prompt \citep{liu2023gevalnlgevaluationusing} defined in Figure \ref{fig:answer_quality_prompt}. We use the small, yet performant OpenAI model \enquote{gpt-4o-mini-2024-07-18} to create the answer, and \enquote{gpt-4.1-mini-2025-04-14} to evaluate the answer, in order to mitigate concerns that the model answering its question would be favorable to its own answer \citep{wataoka2025selfpreferencebiasllmasajudge}. This results in a score between 1-5 for every QA pair. If a QA pair can be fully answered (score of 5), then it is filtered out as being too easy. \textit{Note:} The same approach is used for evaluating the answers of open source models in the Experiments.

\begin{figure}[ht]
\begin{lstlisting}[frame=single, basicstyle=\ttfamily\scriptsize, xleftmargin=0pt, numbers=none]
"""Your task is to answer the QUESTION with the given CONTEXT INFORMATION.

CONTEXT INFORMATION:
---------------------
{context_str}
---------------------

QUESTION: {question}

Given the CONTEXT INFORMATION and not prior knowledge, answer the QUESTION.

Follow the following guideline when answering the QUESTION:
{guideline}
"""
\end{lstlisting}
\caption{Prompt for question answering. \enquote{context\_str} is replaced with a set of sources, \enquote{question} is replaced with the question, and \enquote{guideline} is replaced with its corresponding \textit{answer type} as defined in Figure \ref{fig:guidelines_qa_pair_gen}.}
\label{fig:answer_generation_prompt}
\end{figure}

\begin{figure}[ht]
\begin{lstlisting}[frame=single, basicstyle=\ttfamily\scriptsize, xleftmargin=0pt, numbers=none]
"""You will be given a question, a proposed answer, and a ground truth answer.

Your task is to rate the correctness of the proposed answer. Please make sure you read and understand these instructions carefully.

Evaluation Criteria:
Correctness (1-5) - This evaluates whether the proposed answer is factually correct based on the ground truth. Your task is to determine if the proposed answer is aligned with and entailed by the ground truth answer.


Evaluation Steps:
1. Read the question and ground truth answer: Understand the key facts and details provided in the ground truth answer that are relevant to the question.
2. Compare the proposed answer to the ground truth answer: Check if the proposed answer is factually accurate, consistent, and aligned with the ground truth answer.
3. Assign a correctness score (1-5):
   - 1: The proposed answer is factually incorrect or contradicts the ground truth.
   - 2: The proposed answer contains multiple factual errors or significant inaccuracies.
   - 3: The proposed answer is partially correct but includes some factual errors or omissions.
   - 4: The proposed answer is mostly correct with minor factual deviations.
   - 5: The proposed answer is fully correct and strictly aligns with the ground truth.


Question:
-----
{Question}
-----

Ground Truth Answer:
-----
{Ground_Truth_Answer}
-----

Proposed Answer:
-----
{Proposed_Answer}
-----

Evaluation Form (output ONLY a single score - nothing else):
- Correctness:
"""
\end{lstlisting}
\caption{Prompt for evaluating answer quality. \enquote{Question} and \enquote{Ground\_Truth\_Answer} are the question and answer under investigation. \enquote{Proposed\_Answer} is the newly produced answer that is to be evaluated.}
\label{fig:answer_quality_prompt}
\end{figure}

\section{Filters applied on syn-pdfQA}\label{app:filters_on_syn}
20.4\% of the synthetically generated data is filtered out by the quality filters. This rate is consistent among file types (see also Table \ref{tab:filters_on_syn}). It becomes apparent that most of these QA pairs do not pass the formality filter (see Figure \ref{fig:formality_filter}). Investigating these cases, we see that the filter is rather conservative. The question \enquote{What was the total number of employees in the Group in 2024 according to the Employees table?} is rejected as being inconsistent with respect to the entire document, likely because of the reference to \enquote{the Employees table} where there might be multiple of. We argue that these conservative decisions make the dataset at most more rigorous. In contrast, the internal and external validity of the QA pairs is very high. Only 23.3\% of the errors originate from a lack of internal validity, and 20.9\% from a lack of external validity. Overall, the quality filters seem necessary, but our data generation process already creates mostly valid samples. In turn, we cannot ensure the difficulty with our pipeline. 67.5\% of the error-free QA pairs are fully solvable (G-Eval score of 5, see prompt in Figure \ref{fig:answer_quality_prompt}) by a long context approach with GPT-4o-mini, thereby filtered out. The combination of both filters increases our belief in creating a valid and challenging QA dataset.

\begin{table*}[t]
\tiny
\centering
\begin{tabular}{cccccc}
\hline
                             & Financial reporting & Research articles & Books & Sustainability disclosures & All  \\ \hline
Input QA pairs               & 1922                & 1521              & 1639  & 2573                       & 7655 \\ \hline
Failed validity filter       & 405                 & 279               & 368   & 513                        & 1565 \\
Too easy in difficulty filter & 1149                & 810               & 884   & 1265                       & 4108 \\ \hline
Final dataset                & 368                 & 432               & 387   & 795                        & 1982 \\ \hline
\end{tabular}
\caption{Filters on syn-pdfQA. The table shows the number of data points generated by the synthetic pipeline and filtered by quality and difficulty filters.}
\label{tab:filters_on_syn}
\end{table*}


\section{Filters applied on real-pdfQA}\label{app:filters_on_real}
For \textit{real-pdfQA}, we do not have access to the raw data (i.e., a perfectly parsed state). However, we obtain QA pairs with sources, meaning our formality and internal validity filter can be applied as before. For the external validity filter, we use the parsed PDFs by \textit{PyMUPDF} and transform them into paragraphs using the \textit{sentenceSplitter()} function by LlamaIndex\footnote{\url{https://developers.llamaindex.ai/python/framework-api-reference/node_parsers/sentence_splitter/}} with a chunk size of 400 and 0 chunk overlap. With this method, we can apply all quality filters on the QA pairs. However, this means there might be parsing errors that make our external validity filters more conservative. Again, we argue that filtering too many correct QA pairs out is not a major concern, as long as enough valid pairs are preserved.

As a result, the quality filters eliminate 58\% of the QA pairs (see also Table \ref{tab:filters_on_real}). We observe that a large amount of QA pairs do not pass the formality checks (68.5\% of the errors), mostly because questions are not fully answerable with respect to the whole document. For instance, there are many questions like \enquote{What are the types of audit fees in the table?}. This likely originates from the creation process of these datasets, that intended to have QA pairs on a selected set of sources, not considering validity with respect to the whole document. This is also reflected in a high rate of errors in the external validity. Out of all quality errors, 66.3\% are tagged as lacking external validity (being wrong with respect to the whole document). 58\% of the error cases are tagged as lacking internal validity (being wrong with respect to the relevant sources).

We refrain from making strong claims about original QA dataset quality, as our task (QA on the whole document) differs from some of the original task descriptions and the conservativeness of our filters, which may increase the number of QA pairs tagged as erroneous. This is not the focus of this work, and other projects outline errors more cleanly \citep[e.g.,][]{liu2024ecbdevidencecenteredbenchmarkdesign, klie2022annotationerrordetectionanalyzing, calamai-etal-2025-benchmarking}. As outlined, for us, a conservative filter is not a problem. However, these results signal the usefulness of applying filters to existing human-annotated datasets, even if the task descriptions are highly related. Similarly, we find value in applying the difficulty filter because it eliminates another 78\% of the quality-filtered QA pairs.

\section{Hand-annotation of filtered QA pairs}\label{app:hand_annotation}
We have two major filtering steps: quality filtering and validity filtering. Our quality filtering steps are conservative and filter out a dominant mass of erroneous QA pairs. With our difficulty filter, we sample for QA pairs that are not easy.  A priori, it is not clear whether we may have QA pairs where the gold answer is just wrong, and that is why the model may fail. Thus, we hand-annotate filtered QA pairs to mitigate these concerns.

For \textit{syn-pdfQA}, we apply high scrutiny because, at this stage, only automated filters have been in place. For this reason, we sample 200 filtered QA pairs, 50 random samples of each file type, and annotate them with two annotators. We find that 88\% of the QA pairs are entirely correct with an annotator agreement of 93\% (Cohen's kappa: 53\%; error cases often originate from subjective differences in judgements, as usual in QA annotation tasks). This way, we can mitigate concerns that our pipeline would sample for erroneous QA pairs.

For \textit{real-pdfQA}, we could assume a high quality because all QA pairs are already human-annotated. However, as prior work and our pipeline show, we could also have errors in these datasets. This is why we also randomly sample 100 QA pairs from our final dataset and hand-annotate them with one annotator. We find that 91\% of the QA pairs are indeed entirely correct.

\begin{table*}[h]
\centering
\tiny
\begin{tabular}{cccccclllll}
\hline
                             & FinQA & FinanceBench & ClimateFinanceBench & ClimRetrieve & FeTaQA & NaturalQuestions & PaperTab & PaperText & Tat-QA & All   \\ \hline
Input QA pairs               & 4392  & 148          & 300                 & 167          & 594    & 1410             & 300      & 1068      & 14487  & 22866 \\ \hline
Failed validity filter       & 3045  & 47           & 141                 & 55           & 224    & 1147             & 255      & 766       & 7611   & 13291 \\
Too easy in dificulty filter & 954   & 88           & 91                  & 91           & 301    & 245              & 26       & 231       & 5507   & 7534  \\ \hline
Final dataset                & 393   & 13           & 68                  & 21           & 69     & 18               & 19       & 71        & 1369   & 2041  \\ \hline
\end{tabular}
\caption{Filters on real-pdfQA. The table shows the number of data points taken from the original datasets and filtered by quality and difficulty filters.}
\label{tab:filters_on_real}
\end{table*}


\section{Details on the experimental setup}\label{app:details_on_model_setup}
We choose five open source LLMs to run our experiments: gpt-oss-120b and gpt-oss-20b \citep{openai2025gptoss120bgptoss20bmodel}, qwen3-next \citep{qwen3}, gemma3-27b \citep{gemmateam2025gemma3technicalreport}, and llama3-8b \citep{grattafiori2024llama3herdmodels}. We choose these models because they all have a context size of 128k tokens, making them comparable on the same datapoints. The document in \textit{pdfQA} can be very long, and the parsed version with PyMUPDF is commonly larger than 128k tokens. This means we reduce our datasets from 1982 to 970 in \textit{syn-PDFQA} and 2041 to 1824 in \textit{real-pdfQA}. We argue that this leads at most to an overestimation of the model performance, given that longer documents are more difficult (see also Table \ref{tab:full_result_on_syn}). We evaluate the answer with \enquote{gpt-4.1-mini-2025-04-14} and the prompt in Figure \ref{fig:answer_quality_prompt}, giving us scores in between 1-5.

\section{Results on \textit{syn-pdfQA}}
Table \ref{tab:full_result_on_syn} shows the entire set of results on \textit{syn-pdfQA}, differentiated by all possible complexity dimensions.

\begin{table*}[h]
\centering
\resizebox{\textwidth}{!}{
\begin{tabular}{llccccc}
\toprule
\textbf{Complexity Dimension} & \textbf{Subdimension} 
& \textbf{GPT-OSS-120B} 
& \textbf{GPT-OSS-20B} 
& \textbf{Qwen3-Next} 
& \textbf{gemma3-27b*} 
& \textbf{llama3-8b} \\
\midrule
overall 
& -- 
& 4.399 & 4.047 & 4.062 & 3.726 & 3.498 \\
\midrule
\multirow{4}{*}{file type} 
& financial reporting 
& 4.266 & 3.781 & 3.863 & 3.626 & 3.288 \\
& research article 
& 4.625 & 4.455 & 4.335 & 4.025 & 3.871 \\
& sustainability disclosure 
& 4.104 & 3.586 & 3.780 & 3.241 & 3.010 \\
& book 
& 4.307 & 3.840 & 4.003 & 3.235 & 3.246 \\
\midrule
\multirow{6}{*}{answer type} 
& one-sentence-answer 
& 4.358 & 4.105 & 3.952 & 3.655 & 3.552 \\
& open-ended-question-long 
& 4.881 & 4.300 & 4.573 & 4.013 & 3.461 \\
& open-ended-question-short 
& 4.552 & 4.214 & 4.049 & 3.785 & 3.533 \\
& value-question 
& 4.210 & 3.789 & 4.001 & 3.540 & 3.439 \\
& word-answer 
& 3.721 & 3.497 & 3.505 & 3.200 & 3.071 \\
& yes-no-question 
& 4.528 & 4.221 & 4.377 & 4.286 & 4.137 \\
\midrule
\multirow{2}{*}{reasoning} 
& reasoning 
& 4.323 & 3.974 & 3.964 & 3.593 & 3.363 \\
& replicate 
& 4.479 & 4.125 & 4.165 & 3.867 & 3.641 \\
\midrule
\multirow{3}{*}{question difficulty} 
& complex 
& 4.239 & 3.807 & 3.789 & 3.426 & 3.216 \\
& medium 
& 4.317 & 3.994 & 3.962 & 3.491 & 3.200 \\
& simple 
& 4.465 & 4.128 & 4.163 & 3.872 & 3.656 \\
\midrule
\multirow{3}{*}{modalities} 
& multimodal 
& 4.495 & 4.174 & 4.097 & 3.821 & 3.446 \\
& table 
& 4.177 & 3.707 & 3.808 & 3.411 & 3.123 \\
& text 
& 4.538 & 4.271 & 4.258 & 3.944 & 3.840 \\
\midrule
\multirow{4}{*}{sources position} 
& in the first 25\% 
& 4.375 & 4.032 & 4.055 & 3.735 & 3.459 \\
& in 25-50\% 
& 4.540 & 4.132 & 4.172 & 3.727 & 3.524 \\
& in 50-75\% 
& 4.380 & 4.007 & 4.001 & 3.622 & 3.424 \\
& in 75-100\% 
& 4.303 & 4.018 & 4.016 & 3.810 & 3.586 \\
\midrule
\multirow{2}{*}{\#sources above median} 
& 0 
& 4.295 & 3.906 & 3.957 & 3.627 & 3.460 \\
& 1 
& 4.550 & 4.253 & 4.215 & 3.871 & 3.552 \\
\midrule
\multirow{2}{*}{file length above median} 
& 0 
& 4.560 & 4.352 & 4.271 & 4.001 & 3.788 \\
& 1 
& 4.234 & 3.738 & 3.849 & 3.447 & 3.203 \\
\midrule
\multirow{2}{*}{source spread above median} 
& 0 
& 4.438 & 4.090 & 4.096 & 3.809 & 3.533 \\
& 1 
& 4.359 & 4.005 & 4.027 & 3.643 & 3.462 \\
\bottomrule
\end{tabular}
}
\caption{Model performance per complexity dimension on \textit{syn-pdfQA}. Answers are produced with the parsed full document as a context inputted in the prompt template in Figure \ref{fig:prompt_qa_pair_gen}. The performance is measured with the G-Eval correctness prompt in Figure \ref{fig:answer_quality_prompt}.\\ *The gemma3 tokenizer processes the documents differently, eliminating some datapoints, ultimately rendering the results less comparable.}
\label{tab:full_result_on_syn}
\end{table*}

\section{Results on \textit{real-pdfQA}}
Table \ref{tab:full_result_on_real} shows the results on \textit{real-pdfQA}, differentiated by dataset type.

\begin{table*}[h]
\centering
\resizebox{\textwidth}{!}{
\begin{tabular}{llccccc}
\toprule
\textbf{Complexity Dimension} & \textbf{Subdimension} 
& \textbf{GPT-OSS-120B} 
& \textbf{GPT-OSS-20B} 
& \textbf{Qwen3-Next} 
& \textbf{gemma3-27b*} 
& \textbf{llama3-8b} \\
\midrule
overall 
& -- 
& 3.303 & 3.043 & 3.366 & 2.730 & 2.913 \\
\midrule
\multirow{9}{*}{dataset} 
& ClimRetrieve 
& 3.470 & 2.893 & 1.188 & 1.380 & 1.350 \\
& ClimateFinanceBench 
& 2.387 & 2.558 & 2.785 & 2.342 & 2.253 \\
& FinQA 
& 3.613 & 2.960 & 3.455 & 2.511 & 2.673 \\
& FinanceBench 
& 4.106 & 4.110 & 4.353 & 2.615 & 3.066 \\
& FeTaQA
& 3.705 & 3.373 & 3.597 & 3.133 & 3.267 \\
& NaturalQuestions
& 3.635 & 3.961 & 3.569 & 3.406 & 3.146 \\
& PaperTab 
& 3.484 & 3.368 & 3.308 & 3.029 & 2.557 \\
& PaperText
& 3.724 & 3.870 & 3.475 & 3.192 & 2.980 \\
& Tat-QA
& 3.192 & 3.000 & 3.378 & 2.778 & 3.028 \\
\bottomrule
\end{tabular}
}
\caption{Model performance on \textit{real-pdfQA}. nswers are produced with the parsed full document as a context inputted in the prompt template in Figure \ref{fig:prompt_qa_pair_gen}. The performance is measured with the G-Eval correctness prompt in Figure \ref{fig:answer_quality_prompt}.\\ *The gemma3 tokenizer processes the documents differently, eliminating some datapoints, ultimately rendering the results less comparable.}
\label{tab:full_result_on_real}
\end{table*}

\end{document}